\documentclass[letterpaper, 10 pt, conference]{ieeeconf}
\usepackage{times}
\usepackage[pdftex]{graphicx}
\usepackage{}
\usepackage{amsmath,amssymb,amsopn,amstext,amsfonts}
\usepackage{pifont}

\usepackage{cancel}
\usepackage[space]{cite}
\usepackage{pdfsync}
\usepackage{balance}
\usepackage{color}
\usepackage{mathtools}
\usepackage{algpseudocode}
\usepackage{algorithm}
\algnewcommand\algorithmicforeach{\textbf{for each}}
\algdef{S}[FOR]{ForEach}[1]{\algorithmicforeach\ #1\ \algorithmicdo}

\usepackage{bm}

\usepackage{diagbox}
\usepackage{epstopdf}
\usepackage{pifont}
\usepackage{amsmath}
\usepackage{multirow}
\usepackage{url}
\usepackage{verbatim}
\usepackage{footmisc}
\usepackage[linkcolor=black,citecolor=black,urlcolor=black,colorlinks=true]{hyperref}
\usepackage{color}
\usepackage{tabularx}
\usepackage{float}
\graphicspath{{figures/}}
\DeclareGraphicsExtensions{.png,.jpg,.eps,.pdf}
\IEEEoverridecommandlockouts
\begin{document}

\title{\LARGE \bf Avoiding dynamic small obstacles with onboard sensing and computating on aerial robots}
\author{
Fanze Kong$^{1*}$, Wei Xu$^{1*}$
\thanks{$^{*}$These authors have contributed equally to this work.}, Fu Zhang$^{1}$
\thanks{$^{1}$All authors are with Department of Mechanical Engineering, University of Hong Kong. {\tt\small \{kongfz\}@connect.hku.hk,\{xuweii, fuzhang\}@hku.hk}}
}

\maketitle

\begin{abstract}
In practical applications, autonomous quadrotors are still facing significant challenges, such as the detection and avoidance of very small and even dynamic obstacles (e.g., tree branches, power lines). In this paper, we propose a compact, integrated, and fully autonomous quadrotor system, which can fly safely in cluttered environments while avoiding dynamic small obstacles. Our quadrotor platform is equipped with a forward-looking three‐dimensional (3D) light detection and ranging (lidar) sensor to perceive the environment and an onboard embedded computer to perform all the estimation, mapping, and planning tasks. Specifically, the computer estimates the current pose of the UAV, maintains a local map (time-accumulated point clouds KD-Trees), and computes a safe trajectory using kinodynamic A* search to the goal point. The whole perception and planning system can run onboard at 50Hz with careful optimization. Various indoor and outdoor experiments show that the system can avoid dynamic small obstacles (down to 20mm diameter bar) while flying at 2m/s in cluttered environments. Our codes and hardware design are open-sourced on Github\footnote[2]{\url{https://github.com/hku-mars/dyn_small_obs_avoidance.git}}.

\end{abstract}

\IEEEpeerreviewmaketitle

\section{Introduction}

Unmanned aerial vehicles (UAVs) have shown increasing potentials in a variety of applications, such as delivery, inspection, mapping, and search-and-rescue missions. To enable these widespread applications, a fully autonomous UAV that is able to fly in cluttered environments is crucial. One major challenge in achieving this goal is the detection and avoidance of very small and/or dynamic objects such as tree branches, power lines, small pipes, and stair rails that are common in real-world environments.

Despite the plenty of works on autonomous UAVs recently developed, there are still two critical unsolved issues. Most existing works cannot handle dynamic obstacles with fully onboard UAV resources due to the limited rate of sensing and/or planning. Avoiding small obstacles is another challenge to UAVs due to the low resolution of mapping in existing works. 

These two issues are rooted in the limited sensing capabilities available for existing UAVs. The popularly used RGB-D sensors have limited sensing range (a few meters) and resolution, preventing the timely and reliable detection of small obstacles (e.g., pipes with diameter smaller than 20mm). The noisy measurements of RGB-D sensors also require additional probabilistic filtering on each grid occupancy (e.g., OctoMap \cite{hornung_octomap_2013}), which imposes an inherent trade-off between processing time and affordable resolution. Multi-line spinning lidars are another type of sensor that has been used on UAVs. Besides the high-cost, their bulky size, weight ($\sim$1kg), low resolution (e.g., $2^{\circ}$ for 16 lines lidars), and low frame-rate (e.g., 10Hz) have significantly prevented their adoption to UAVs for obstacle avoidance. 

\begin{figure}[t]
\centering
\includegraphics[width=0.47\textwidth]{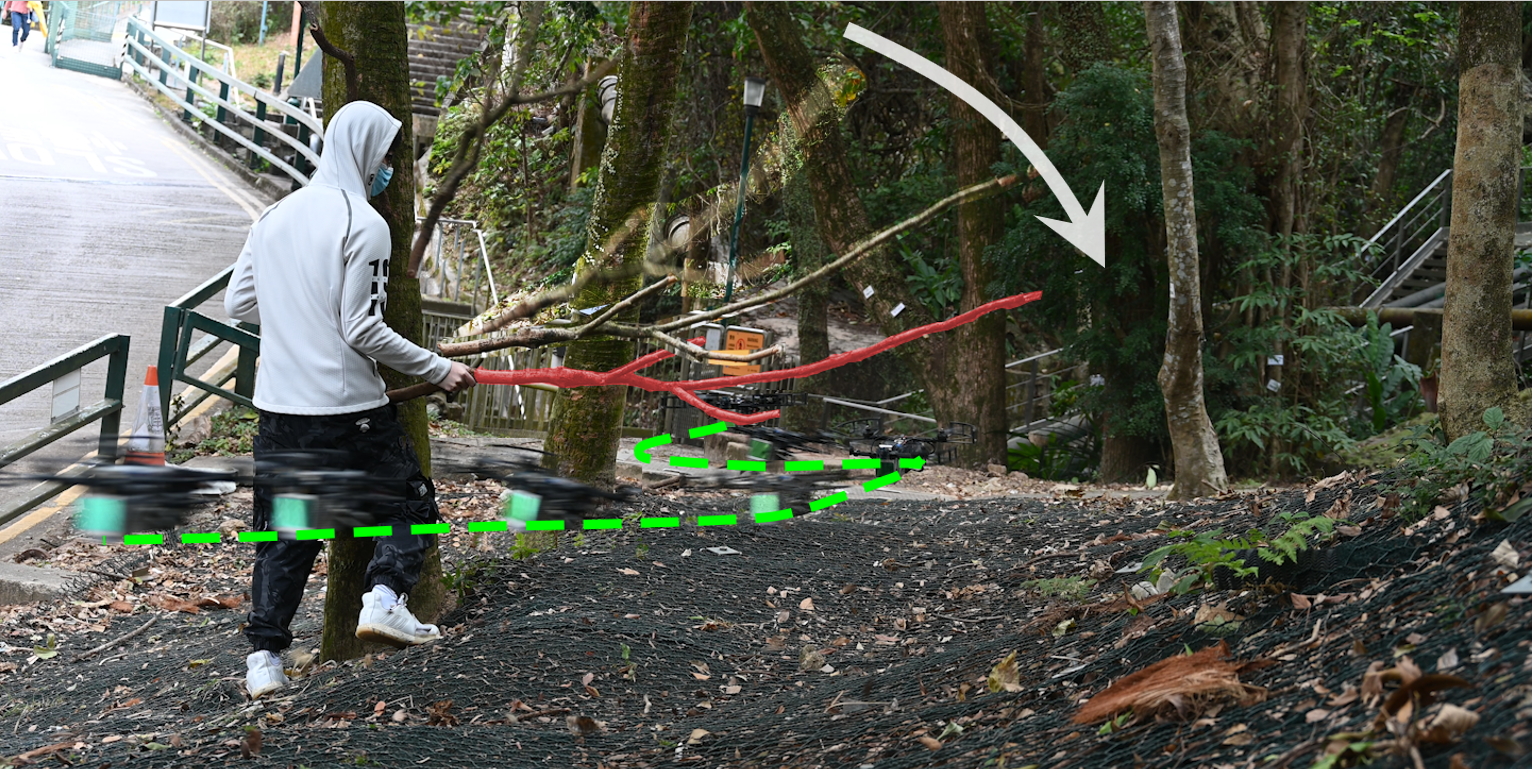}
\caption{Our system is able to detect and avoid dynamic small obstacles such as tree branches. The green dashed line is the path flied by the UAV, the white arrow indicates the movements of the tree branch. The current tree branch is highlighted in red. Video is available at \url{https://youtu.be/pBHbQ_J1Qhc}.}
\label{cover_image}
\vspace{-10mm}
\end{figure}

In this paper, we present an autonomous UAV system aiming to fulfill intelligent flight in cluttered and dynamic environments. The system-level contribution is the proof of feasibility of using only a low-cost, light-weighted, and small form factor 3D lidar and onboard processing to achieve high frequency and safe navigation in unknown cluttered environments. We summarize our contributions as follows:

\begin{enumerate}
\item A complete autonomous lidar-based UAV. It is lightweight (1.8kg in total) and compact (280mm wheelbase) while equipped with onboard sensing (a solid-state lidar: LIVOX AVIA) and computing modules (a DJI Manifold 2-C, with Intel i7 8550U CPU). The system performs full state estimation and high-resolution mapping at 50Hz with onboard computing;
\item A lightweight motion primitives-based motion planner directly running on lidar point-cloud. The motion planner can achieve up to 50Hz replanning rate, on actual onboard systems (as opposed to maximal 30Hz update rate of RGB-D cameras such as Intel Realsense), allowing timely detection and avoidance of dynamic obstacles;
\item The UAV system is able to avoid dynamic small obstacles. Extensive real-world experiments show that our system can reliably detect and avoid both static and dynamic small obstacles (down to 20mm) in various indoor and outdoor environments.
\end{enumerate}

\section{RELATED WORK}

\subsection {Navigation in cluttered environments}
In recent years, the localization and mapping technology of unmanned vehicles have developed rapidly, especially the vision-based and lidar-based methods. For the application of vision-based UAVs, Visual-Inertial Odometry (VIO) is one state-of-art technique, which gives stable odometry by only using camera images and IMU measurement\cite{forster2016svo, bloesch2017iterated, lin_autonomous_2018,bucki_rectangular_2020}. Most existing works of autonomous aerial vehicle system are based on visual sensors and have achieved outstanding performance in static complex environments. 

A standard process of a vision-guided aerial vehicle is to obtain the location from the VIO algorithm and combine it with the depth image from the depth camera to build a map (e.g. Occupancy grid map\cite{hornung_octomap_2013}, Euclidean Signed Distance Field (ESDF) map\cite{oleynikova_voxblox_2017} or point clouds map\cite{liu_robotic_2016}), and then generate and execute a trajectory on the established map. Liu \textit{et al.}\cite{sikang_liu_high_2016} use a uniform
resolution volumetric occupancy grid map to establish a short-range planner for high-speed quadrotor navigation. Oleynikova \textit{et al.}\cite{oleynikova_continuous-time_2016} presents a UAV system using a high rate local replanner based on ESDF map, which can avoid newly-detected obstacles and generate feasible trajectories online in 50ms. Brett \textit{et al.}\cite{lopez_aggressive_2017} proposed a state-based motion primitive planning on current point cloud map and achieved high-speed avoidance. The Nanomap\cite{florence_nanomap_2018} and Mapless planner\cite{ji_mapless-planner_2020} are more elegant, which checks collision not only in the current FoV point cloud, but also in the past FoVs. Nanomap\cite{florence_nanomap_2018} even achieved 10m/s velocity flight in forests considering the pose uncertainty.

Besides vision-based UAV, there are also many works on lidar-based UAV. Liu \textit{et al.}\cite{liu_planning_2017} generate Safe Flight Corridors (SFC) consisting of multiple ellipsoids from point clouds of a 360° lidar and optimize the trajectory in the flight corridor. Mihir \textit{et al.} \cite{dharmadhikari_motion_2020} proposed a method to explore a mine autonomously, which utilizes control-based motion primitives to find future-safe paths in an ESDF map built from lidar data. And Zhang \textit{et al.}\cite{zhang_falco_2020} achieved a 10m/s flight in forests using a set of static motion primitives assuming a constant flying velocity in the lidar frame. 

\subsection {Navigation in dynamic environments}
The approaches mentioned above have a common assumption that the environment is static, so the rate of map updating and replanning is usually low. To avoid dynamic obstacles, it requires both the environment perception (i.e., mapping) and trajectory replanning to be performed at a very high rate. Sanchez \textit{et al.}\cite{sanchez-lopez_real-time_2019} generates a progressive optimal trajectory allowing the UAV to avoid moving people at a very low speed. Allen \textit{et al.}\cite{allen_real-time_2019} use kinodynamic RRT* algorithm to make the UAV avoid a dynamic sword rapidly, but the perception and replanning are both performed by offboard sensors (i.e., Motion capture systems) and computers. Gao \textit{et al.}\cite{gao2020teach} successfully demonstrate the avoidance of dynamic obstacles (e.g., moving people) using a RGB-D camera on a quadrotor UAV, but the system requires a global map saved on an offboard server. Falanga \textit{et al.}\cite{falanga_dynamic_2020} use event camera to fulfill a high speed (10m/s relative speed) objects avoidance, having only 3.5ms overall latency. However, the event cameras cannot have a complete perception of the environment.

\subsection {Navigation considering small obstacles}
In practical applications, it is common for UAVs to encounter small obstacles (e.g., power lines in outdoor, small pipes and/or stair rails in indoor). Existing works are usually based on visual sensors and detect such small objects by reasoning the image. Madaan \textit{et al.}\cite{Madaan-2018-107320} trains a convolution network to segment wires from images and reconstructs them to fulfill the avoidance. Zhang \textit{et al.}\cite{zhang_detecting_2019} proposes a method to detect the power line using convolutional and structured features. These methods are usually not general and costly for UAVs.

Our work considers the three challenges all at once. We emphasize a wholistic system design from sensing, perception, estimation, to planning, instead of focusing on one component alone. Compared with the existing work, our developed system is lightweight, fast, and able to safely navigate in complex indoor and outdoor environments while avoiding dynamic small obstacles. 

\section{SYSTEM OVERVIEW}

\begin{figure}[t!]
\centering
\includegraphics[width=0.5\textwidth]{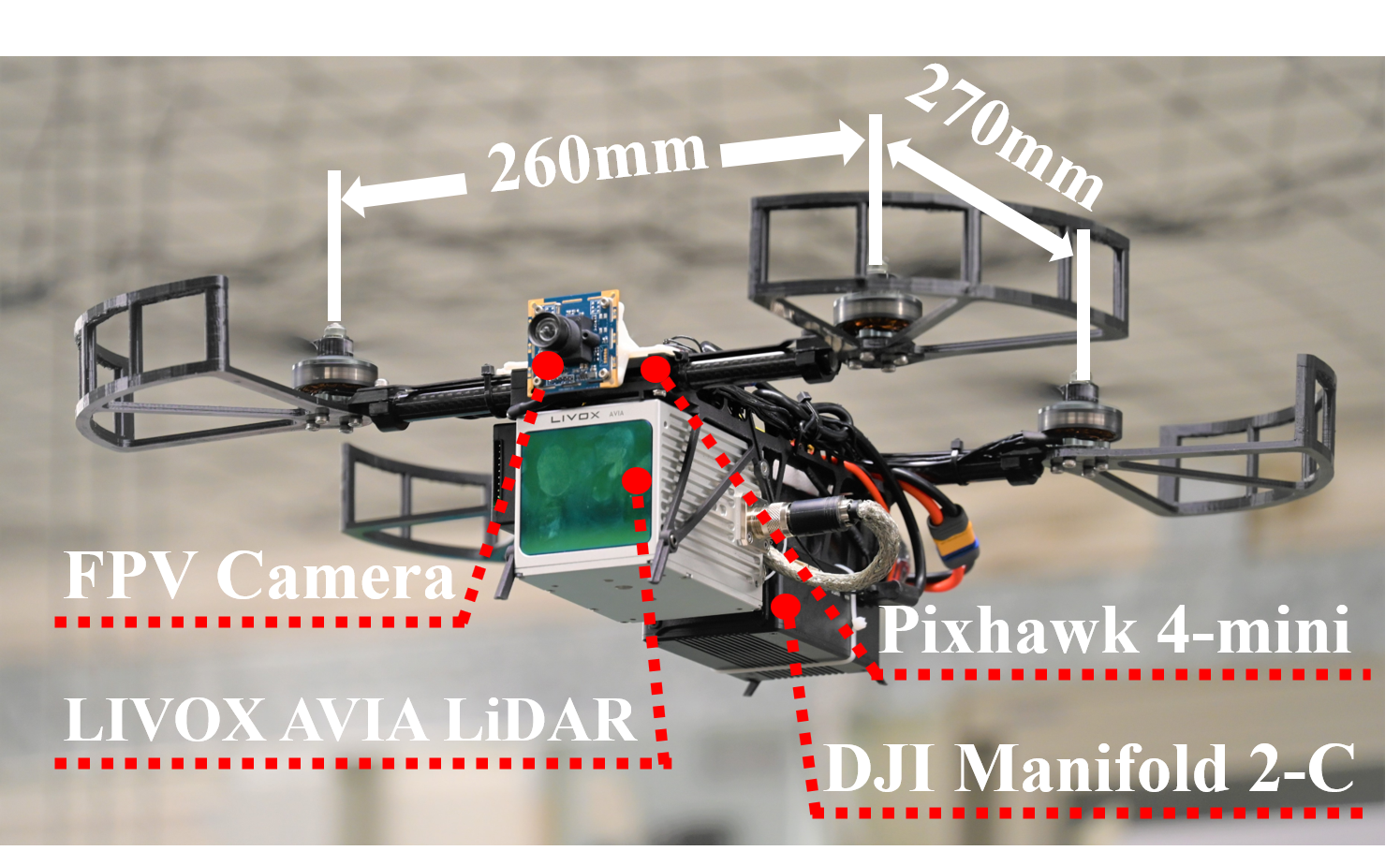}
\caption{Our quadrotor UAV system. The camera is used for visualization only. }
\label{hardwareimage}
\vspace{-8mm}
\end{figure}

In this section, we present the hardware design of our quadrotor testbed aimed at achieving good autonomous flying performance, being an integrated, durable and aggressive UAV, as shown in Fig. \ref{hardwareimage}.

\renewcommand\arraystretch{1.3}
\begin{table}[ht]
\caption{System specifications}
\centering
\begin{tabular}{|p{3cm}|l|}
\hline
Thrust weight ratio         & 3\\ \hline
Weight (lidar included)     & 1.8kg\\ \hline
Size (with propeller)        & 44$cm$x38$cm$x15$cm$ \\ \hline
Size (without propeller)     & 26$cm$x27$cm$x15$cm$\\ \hline
Maximum flight time         & 10min\\ \hline
Battery                     & 5S 3300mah 50C\\ \hline
\end{tabular}
\label{hardware parameters}
\end{table}

\subsection {Fully onboard hardware architecture}

The UAV hardware system consists of a Livox AVIA lidar, a DJI Manifold 2-C onboard computer (i7 8550U CPU, 1.8GHz Basic Frequency, quad-core), and a Pixhawk4-mini flight controller. We also add a monocular camera on the UAV to gain first-person view (FPV) images for better visualization. Note that the camera is not used in the perception system.

Compared with vision-based UAVs, our system only uses a lidar (with a built-in consumer-grade IMU) to navigate the UAV and to sense the environment. The measured point cloud and IMU data are transmitted to the onboard computer, which performs all the state estimation, mapping, and motion (re-) planning (see Section \ref{section:software architecture}). At last, the flight controller Pixhawk4-mini performs the trajectory tracking control. 

Besides the avionic system mentioned above, we also design and optimize our specific quadrotor frame. To ensure that the UAV has enough mobility and small size, we choose T-motor F90 motors and 7042 propellers as the power plant. As shown in Table \ref{hardware parameters}, the final UAV has a moderate weight (1.8kg in total), size (280mm wheelbase), and flight endurance (over 10 mins). With over 3-g thrusts, the UAV is able to perform aggressive maneuvers to quickly respond to dynamic obstacles. 

\subsection{Lidar characteristics}
\begin{figure}[t]
\centering
\includegraphics[width=0.38\textwidth]{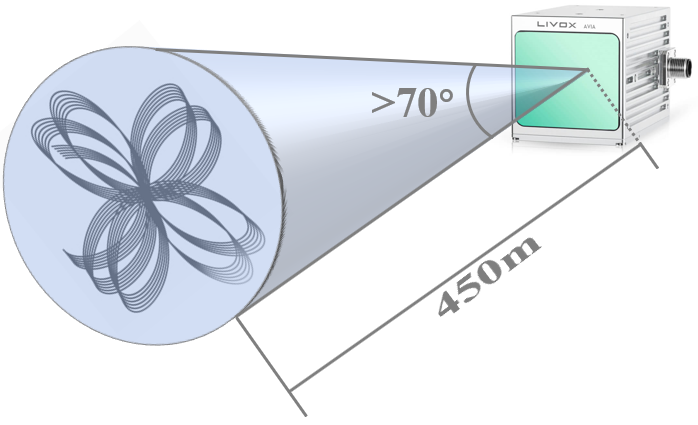}
\caption{Field of view and scanning pattern of livox AVIA lidar. }
\label{lidarfigure}
\end{figure}

\begin{figure}[t!]
\centering
\includegraphics[width=0.45\textwidth]{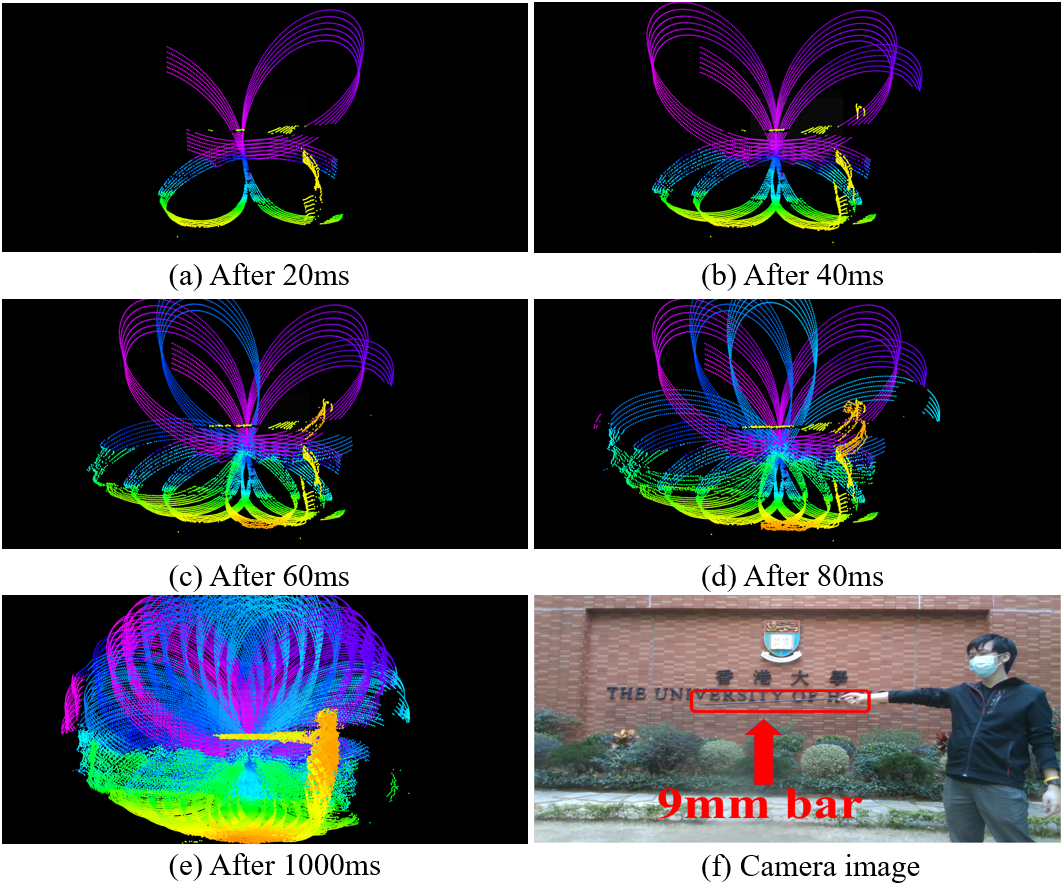}
\caption{Point clouds scanning results of a 9mm diameter bar over different accumulation time. The bar can be detected in the first 20ms data.}
\label{lidarscan_principle}
\end{figure}

Lidar is a type of sensor which emits laser pulses and receives the reflected pulse to gain depth information of the environment. Unlike multi-line spinning lidars used on existing UAVs \cite{liu_planning_2017}, we use a solid-state lidar, the Livox AVIA, which features many advantages that are suitable for aerial applications: (1) Shown in Table \ref{Lidar parameters}, the sensor has a weight and size that can be adopted to a small-scale UAV; (2) the sensor has a very long detection range, allowing obstacle detection and avoidance at high-speed flight; (3) the sensor has an extremely low false alarm rate, producing only three noise points per one million points. This could eliminate the time-consuming ray-casting occupancy filtering as in Octomap \cite{hornung_octomap_2013} and ESDF \cite{oleynikova_voxblox_2017}; (4) the sensor can output point clouds data at a very high rate (up to 2500Hz), as opposed to the 10$\sim$20Hz output rates of multi-line spinning lidars. This allows timely detection of dynamic obstacles; (5) the sensor uses 6 laser heads to scan the front area simultaneously in a rosellete pattern, forming a circular FoV (see Fig. \ref{lidarfigure}). Since it covers the entire FoV very quickly, the sensor is very suitable for detecting line-shaped small objects. Furthermore, the sensor uses a non-repetitive scanning \cite{livox} that significantly boosts the resolution even being static. An example of the sensor measurements for a 9mm bar is seen in Fig. \ref{lidarscan_principle}, it can be seen that the bar is detected only after 20ms and the resolution increases over the accumulation time. 

\renewcommand\arraystretch{1.3}
\begin{table}[ht]
\caption{Lidar Parameters}
\centering
\begin{tabular}{|p{3.5cm}|l|}
\hline
Maximum Detection Range     & 450m\\ \hline
FoV                         & $70.4^{\circ} \times 77.2^{\circ}$ \\ \hline
Point Rate                  & 240,000points/sec\\ \hline
Maximum Frame Rate          & 2500Hz\\ \hline
Build-in IMU Rate           & 200Hz\\ \hline
Range Precision             & 2cm\\ \hline
False Alarm Ratio           & \textless0.0003$\%$\\ \hline
Weight                      & 498g\\ \hline
Size                        & 76mm$\times$65mm$\times$91mm\\ \hline
\end{tabular}
\label{Lidar parameters}
\end{table}

\section{Integrated Onboard Perception And Motion Planning}
\label{section:software architecture}

\begin{figure}[t]
\centering
\includegraphics[width=0.50\textwidth]{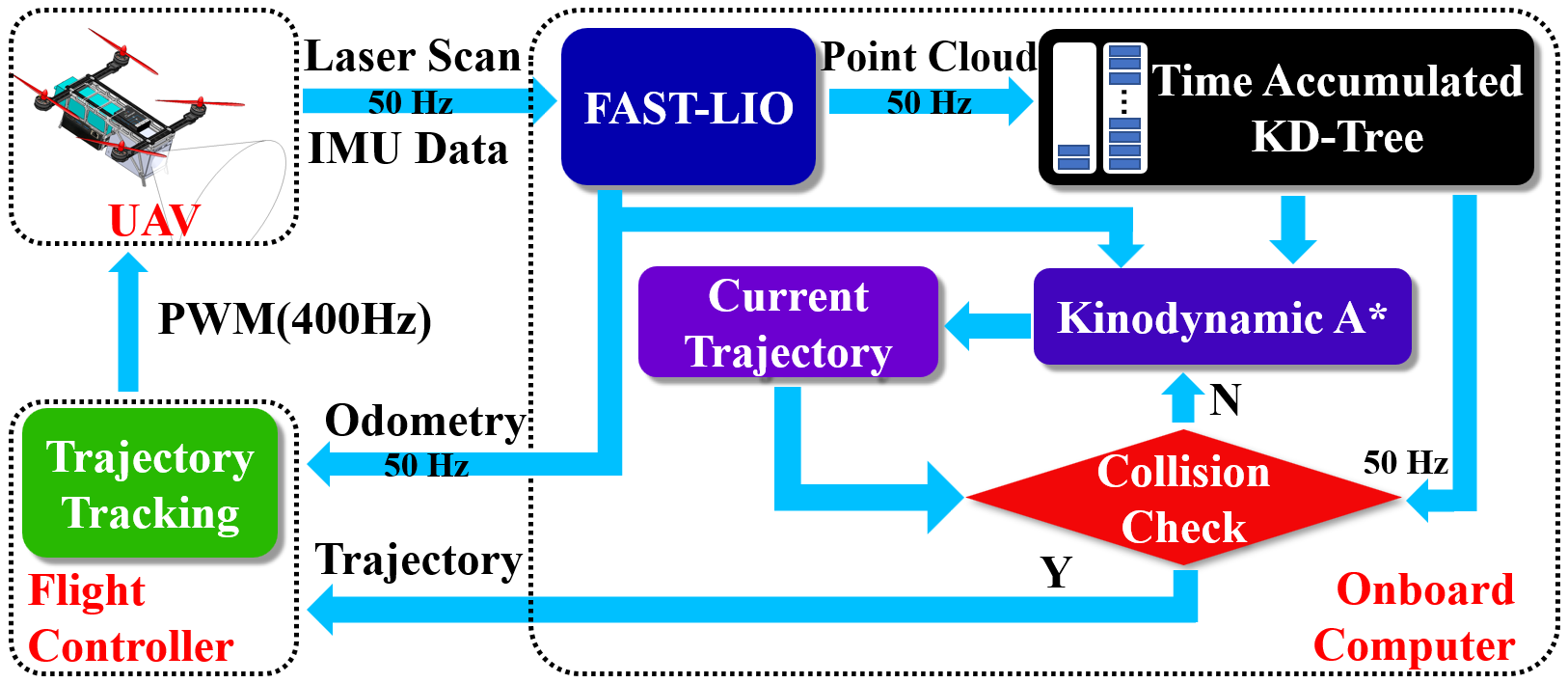}
\caption{The software architecture of our UAV system. Perception and planning are all running on the onboard computer.}
\label{framework_figure}
\end{figure}

In this section, we detail the software and algorithm design. The overview of the system workflow is seen in Fig. \ref{framework_figure}. A typical autonomous UAV consists of three functional parts: navigation, planning, and control. In our system, navigation and planning modules are both running on the onboard computer, and the tracking control is running on the onboard flight controller Pixhawk4-mini. For the navigation task, we use FAST-LIO lidar SLAM (Simultaneous Localization And Mapping)\cite{xu_fast-lio_2020} algorithm to compute the location of the UAV and build a point cloud map of the environment, both at 50Hz. After receiving a new scan of point cloud and IMU data from the lidar, FAST-LIO estimates the current UAV state in a tightly-coupled iterated Kalman filter. The estimated state is, in turn, used to project the new scan of point cloud to the world frame. The new scan of point cloud is then added to a local map organized in two time-accumulated KD-Trees, detailed in Section \ref{section:time_accumulated_kdtree}. The updated local map triggers a collision check on the current trajectory under tracking. If any collision exists, a re-planning is triggered, and a new safe trajectory is generated by the kinodynamic A* search, as detailed in Section \ref{sec:Astar}, to replace the current trajectory. Otherwise, the current trajectory along with the state estimation from FAST-LIO are sent out to the controller for tracking control. The trajectory tracking controller is a cascaded PID controller implemented on Pixhawk4-Mini, and we tune the parameters appropriately beforehand. Furthermore, we carefully optimize all the software modules to enhance their real-time performance. For example, we replace the KD-tree structure with an incremental KD-tree structure \cite{cai_ikd-tree_2021} in FAST-LIO to speed up its computation. 

\subsection {Time-accumulated KD-Tree}
\label{section:time_accumulated_kdtree}

Most existing works rely on occupancy grid map\cite{sikang_liu_high_2016, gao2020teach, oleynikova_continuous-time_2016}, where the space is discretized into small grids. To filter out sensor noise, ray-casting is usually conducted to estimate the occupancy probability of each grid along the ray of a point measurement (e.g., OctoMap \cite{hornung_octomap_2013}). Euclidean Signed Distance Field (ESDF) map\cite{oleynikova_voxblox_2017, han2019fiesta} builds on top of the occupancy grid map and further maintains the signed distance of each grid to its nearest occupied grids. Such distance provides the gradient information allowing the optimization of a smooth trajectory. However, it takes additional time to compute the distance information.

Due to the very long measuring range of our lidar (up to 450m as in Table \ref{Lidar parameters}), ray-casting is very time-consuming, especially when the grid size is small to accommodate small objects. On the other hand, the lidar has a very low noise ratio (lower than 0.0003$\%$) that makes the ray-casting used in OctoMap unnecessary. In this paper, we directly plan on the raw point clouds received from the estimation and mapping module. To enable efficient collision check \cite{lopez_aggressive_2017}, we organize the raw point clouds into a KD-tree structure.

Considering all historic points in the collision check leads to over conservative motion planning since the space travelled by dynamic objects will be mistakenly considered as occupied \cite{gao_flying_2019}. Moreover, maintaining all points in a single KD-tree is time-consuming due to the large number.  On the other hand, using points in the current scan only \cite{lopez_aggressive_2017} will lead to incomplete coverage of the FoV and miss potential obstacles on the trajectory.

\begin{figure}[t]
\centering
\includegraphics[width=0.91\linewidth]{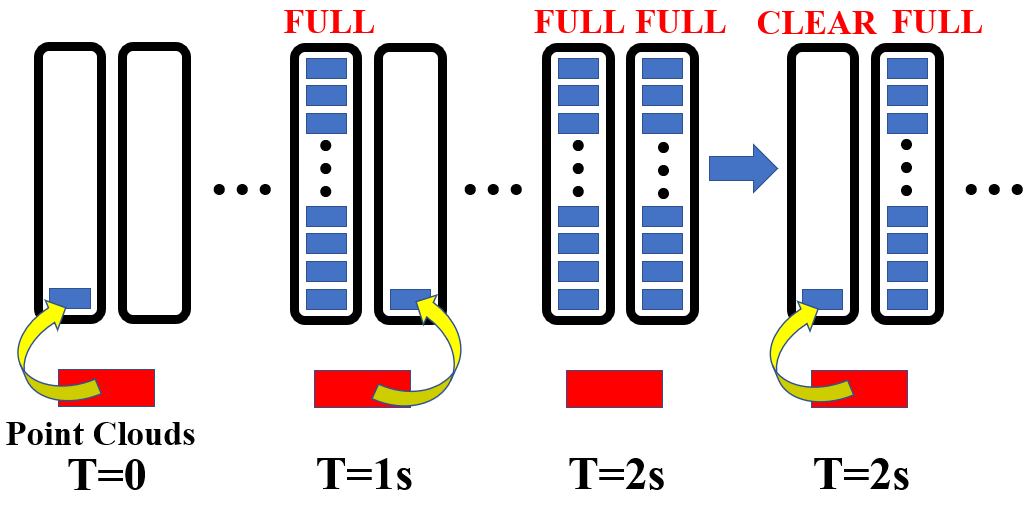}
\caption{Update of the two time-accumulated KD-Trees with a new scan of points.}
\label{accumulatekdtrees}
\vspace{-9mm}
\end{figure}

We propose to use a temporal local map. That is, only the most recent point cloud is used for collision check. We accumulate the point cloud over a certain time, called the accumulation time. For the livox AVIA lidar, the accumulation time is set to one second, which produces sufficiently high-resolution point clouds (see Fig. \ref{lidarscan_principle}). We further downsample the raw point clouds at a prescribed resolution (e.g., 10cm) to lower the computation. Note that the down-sampling resolution, unlike the grid size in Octomap, does not affect the detection of objects smaller than the resolution, but may only inflate the obstacle space slightly. 

The local map is organized into two KD-Trees, one static and one dynamic growing (see Fig. \ref{accumulatekdtrees}), each KD-Tree stores up to the accumulation time point clouds data. The update of each KD-tree is detailed in Algorithm \ref{alg:timeaccumulatedKDtrees}, where $H = 50$ denotes the maximum number of lidar scans stored in one KD-Tree, and $N = 2$ denotes the number of KD-Trees. In the beginning, the two KD-Trees are both empty. Once a new scan of points arrives, the KD-Tree that is still not full is obtained  (Line 1 - 6). Then, the new scan of points is added to the point cloud already on the Tree (Line 11). The updated point clouds are down sampled at the prescribed resolution (Line 13) and used to build up a new KD-Tree replacing the existing one (Line 14). The above process repeats until the current KD-Trees is full (Line 7), where the new scan is accumulated from scratch (Line 8-9) and saved to the other KD-Tree (this will override the existing point cloud already on the other KD-Tree). 

\begin{algorithm}[ht]
  \caption{Time-accumulated KD-Tree}
    \hspace*{0.02in} {\bf Input: }
    Point clouds $NewPoints$ in the new scan \\
    \hspace*{0.02in} {\bf Output:}
    Two KD-Trees $KDTree[2]$
    \\
    \hspace*{0.02in} {\bf Parameters:}
    $H = 50 \quad N = 2$
    
    \begin{algorithmic}[1]
    \If{$ScanInputNum \geq H * N$}
        \State $ScanInputNum = 0;$
        \State $TreeInputNum = 0;$
    \Else
        \State{$TreeInputNum\!=\! \lfloor ScanInputNum/H \rfloor;$}
    \EndIf
    \If{$ScanInputNum \bmod H = 0$}
        \State $CloudAccumulate.Clear();$
        \State $CloudAccumulate = NewPoints;$
    \Else
        \State $CloudAccumulate.add(NewPoints);$
    \EndIf
    \State $CloudFilterd = VoxelGridFilter(CloudAccumulate);$
    \State $KDTree[TreeInputNum].build(CloudFilterd);$
    \State $ScanInputNum = ScanInputNum + 1;$

  \end{algorithmic}
  \label{alg:timeaccumulatedKDtrees}
\end{algorithm}

The two KD-Trees save point cloud up to twice the accumulation time, which is sufficient to cover the lidar FoV with great details. When performing collision check, both KD-Tree should be used: the static one provides information of the environments while the dynamic one timely detects dynamic obstacles at the lidar scan rate (i.e., 50Hz). 

{\it Remark:} when adding a new scan of point cloud, the entire KD-Tree is re-built from scratch (Line 14, Algorithm \ref{alg:timeaccumulatedKDtrees}) by using the standard PCL library \cite{muja2009flann}. This re-building could be very inefficient when the number of new points is significantly less than existing points on the KD-tree. In this case, incremental KD-Tree \cite{cai_ikd-tree_2021} could be used, which saves much building time by only updating and re-balancing the KD-Tree partially.

\subsection {Kinodynamic A* search}
\label{sec:Astar}

Once a collision is found on the current trajectory under tracking (e.g., due to dynamic or newly detected obstacles), the re-planning module is triggered to plan a safe trajectory based on the most recent local map saved in the two KD-Trees and on the current UAV states estimated from FAST-LIO. To avoid dynamic obstacles, the re-planning module should be lightweight and can run at a high frequency. In our system, we choose the motion primitive-based method, which has been successfully used in high-speed planning\cite{lopez_aggressive_2017} and dynamic objects avoidance\cite{falanga_dynamic_2020}. 

We directly adopt the Kinodynamic A* search algorithm presented in \cite{liu2017search}, where a graph is constructed and searched via A*. The graph is initialized with the starting UAV state (e.g., position, velocity, {\it etc.}), where we discretize the control space into a certain number of discrete actions. For each action, we integrate it over a certain time (i.e., $T_s$) and obtain a dynamically feasible trajectory (called {\it motion primitive}). Collision check is performed on each motion primitive using points in the local map (i.e., the two time-accumulated KD-Trees). If the motion primitive is collision-free and satisfies kinodynamic constraints, its end state is added to the graph. This process is called {\it expansion} and is repeated until the target point is reached.  Similar to \cite{liu2017search}, we use {\it analytical expansion} to speed up the search process. 

Since a quadrotor UAV is differential flat \cite{mellinger2011minimum}, the state and actual control can be uniquely determined from the trajectory and its high-order derivatives (the yaw angle is always aligned with the velocity direction), this enables us to use the high-order derivatives (e.g., acceleration, jerk, snap) as the virtual control in Kinodynamic A* search. As shown in \cite{liu2018search}, planning in higher-dimension space such as jerk is usually very time consuming, we choose acceleration as the control action as in \cite{zhou_robust_2019}. More specifically, we discretize the acceleration space of each direction into three actions $\left \{ -a_{\max }, 0, a_{\max } \right\}$, leading to a total number of 27 motion primitives when expanding a node. More implementation details can be referred to \cite{zhou_robust_2019}. 

\section{Results}
\begin{figure*}[t]
\centering
\includegraphics[width=0.99\textwidth]{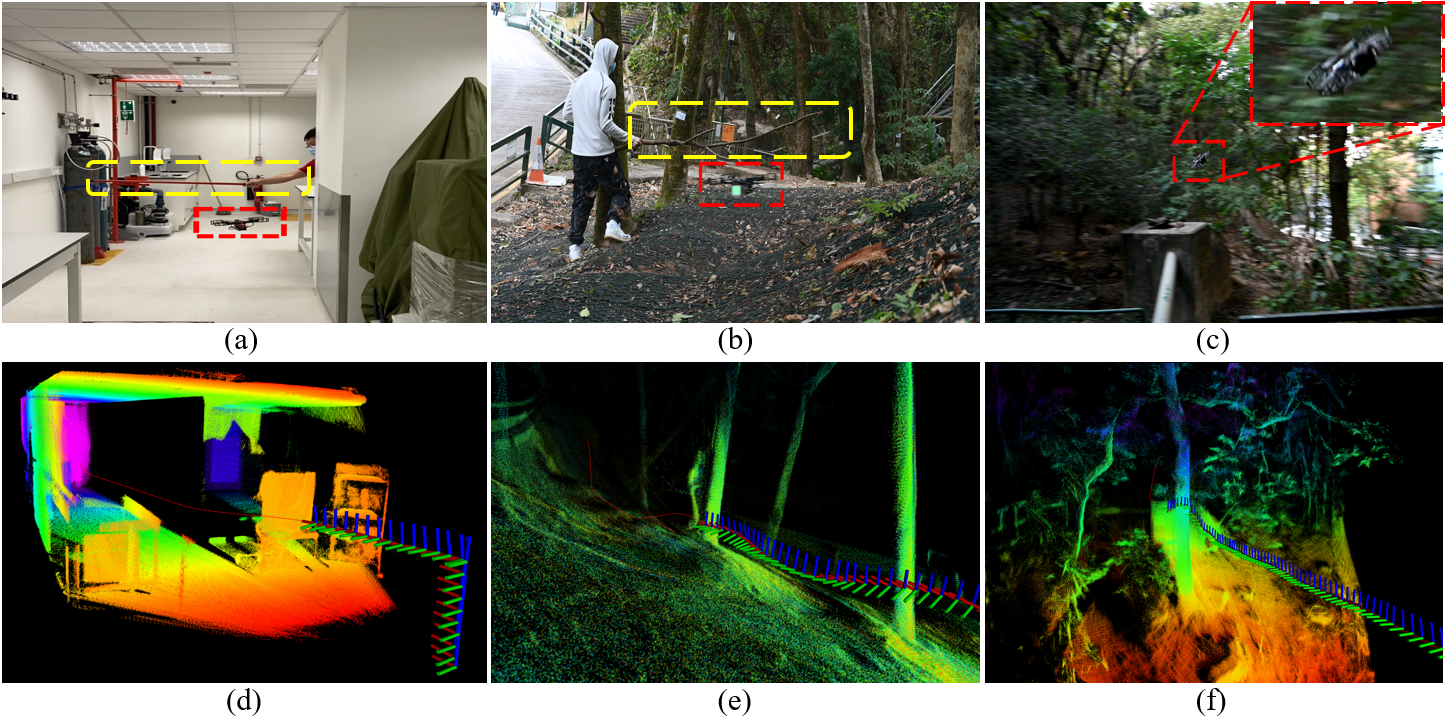}
\caption{We test our developed system in both indoor and outdoor environments: (a) an indoor office corridor with a narrow bar of diameter 20mm being intentionally raised to block the UAV flight path; (b) outdoor forest with a tree branch of diameter {10-20mm} being intentionally lowered to block the UAV flight path; (c) a hillside with cluttered tree crowns. (d-f), the point cloud map built in real-time as the UAV flies in the three tested environments. Notice that (e) is the scene in (b) but rendered at an apposite view angle. }
\label{thirdpersonviewiamge}
\end{figure*}

We present several real-world experiments to validate the hardware and software of our UAV system, indoor office corridor, outdoor forests, and cluttered tree crowns (see Fig. \ref{thirdpersonviewiamge}). All the environments are natural except the possible dynamic small obstacles (e.g., narrow bar, tree branches) intentionally added to the scene. In all experiments, the parameters of the system are set as follows according to the actual situation: the down-sampling resolution is set to 10cm when building the two time-accumulated KD-Trees. The safety clearance between UAV and obstacles is set to 45cm considering the actual size of the UAV and the possible obstacle shrink caused by point cloud down-sampling. The kinodynamic limits of the motion planning modules for velocity is set to 2m/s, $a_{\max} = 2$m/s\textsuperscript{2}, and $T_s = 0.6$s. The snapshots of experiments are shown in Fig. \ref{thirdpersonviewiamge} (a-c). Due to the use of lidar, a byproduct of our system is a high-resolution high-accuracy 3D map of the environments built in real-time as shown in Fig. \ref{thirdpersonviewiamge} (d-f). We supply a video attachment (also available at \url{https://youtu.be/pBHbQ_J1Qhc}) that better visualize the flight performance of our system in all these three environments.

\begin{figure}[ht!]
\centering
\includegraphics[width=\linewidth]{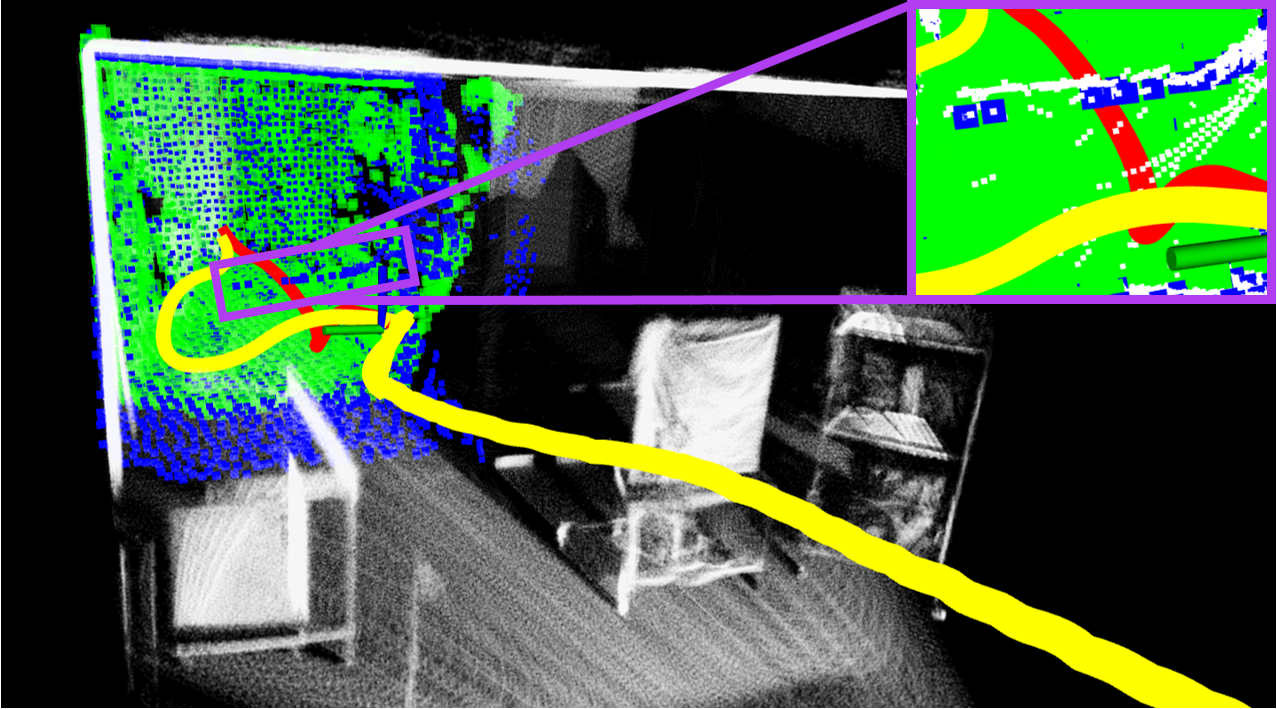}
\caption{Avoiding a dynamic 20mm diameter bar in an indoor environment. The white point clouds represent all points accumulated so far, the blue and green point clouds represent points in the first and second KD-tree of our local map, respectively, used for re-planning. The yellow curve is the current trajectory and the red one is the re-planned one after detecting the new bar position. The upper right corner is a zoomed view of the points collected on the moving bar. }
\label{indoor_environment_figure}
\end{figure}

\subsection{Indoor experiments with dynamic and static obstacles}
Instead of flying in a man-made indoor environment\cite{zhou_robust_2019,lopez_aggressive_2017}, we choose a natural office corridor and no changes have been made to the site, as shown in Fig. \ref{thirdpersonviewiamge}(a). The UAV can autonomously fly to a 9 meters away goal point while avoiding both the static and dynamic obstacles with a maximum velocity of 2m/s. During the flight, we suddenly raise a narrow-bar shape obstacle (diameter 20mm) and hold it to the UAV flight path. However, our system can detect this dynamic small obstacle and avoid it in a very short time. As shown in Fig. \ref{indoor_environment_figure}, a smooth trajectory (the yellow curve) is first generated to the target point. With the narrow-bar raising, the trajectory is detected to have collision with the updated environment, and the planning module starts replanning. After several times of replanning, a final safe trajectory (the red curve) that passes under the bar is planned. In the upper right corner of Fig. \ref{indoor_environment_figure}, we show the zoomed-in view of the narrow bar where the blue points represent the points in time-accumulated KD-Trees used for replanning and white points are points in the past. It is seen that the bar movement profile is clearly captured by the LiDAR points. By using a temporal local map, the re-planner is able to utilize the space where dynamic obstacles swept over (indicated by white points in the upper right corner) to avoid the bar at its current position (indicated by the blue points in the upper right corner). 

\subsection{Outdoor experiments with natural and dynamic obstacles}
\begin{figure}[t]
\centering
\includegraphics[width=\linewidth]{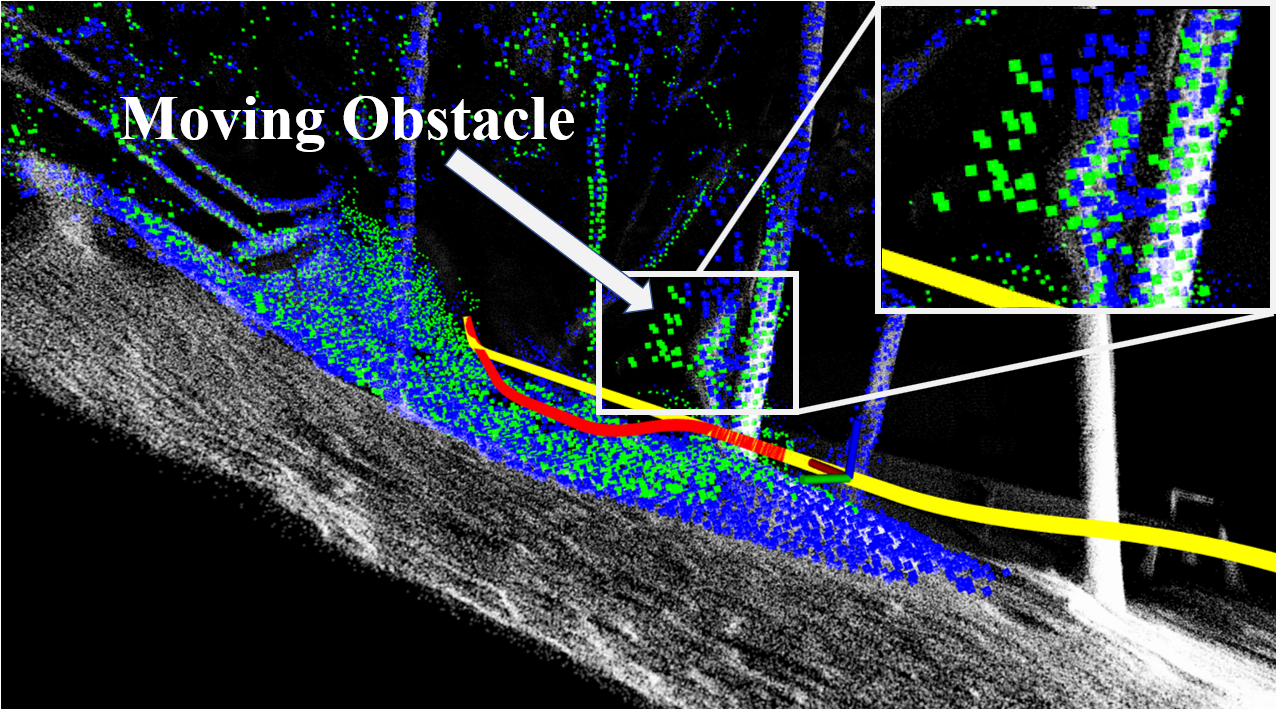}
\caption{Avoiding a dynamic tree branch in outdoor environment. The meanings of point clouds and trajectories are the same as those in Fig. \ref{indoor_environment_figure}.}
\label{moving_obstacle_figure}
\vspace{-10mm}
\end{figure}

In order to validate the system in more natural and complex environments, we also carry out experiments in a forest. In the first scene, the quadrotor passes several trees and a dynamic branch (see Fig. \ref{thirdpersonviewiamge}(b)) to reach the target point 15m away. During the flight, we suddenly swept down a tree branch (diameter 10-20mm) and hold it on the UAV flight path (see Fig. \ref{cover_image}). Fig. \ref{moving_obstacle_figure} shows the moment the re-planning is triggered. It can be seen that the tree branch movement profile is clearly seen (the zoomed view in the upper right corner). Once the tree branch falls within the safety clearance of the current trajectory under tracking (the yellow trajectory), the re-planning is triggered to produce a new trajectory (the red trajectory) that avoids the tree branch. Unlike the indoor experiment, all points on the tree branch are on the KD-Tree. This is because the tree branch falls within the safety clearance, which triggers the re-planning, within twice the accumulation time. 

\begin{figure}[t]
\centering
\includegraphics[width=\linewidth]{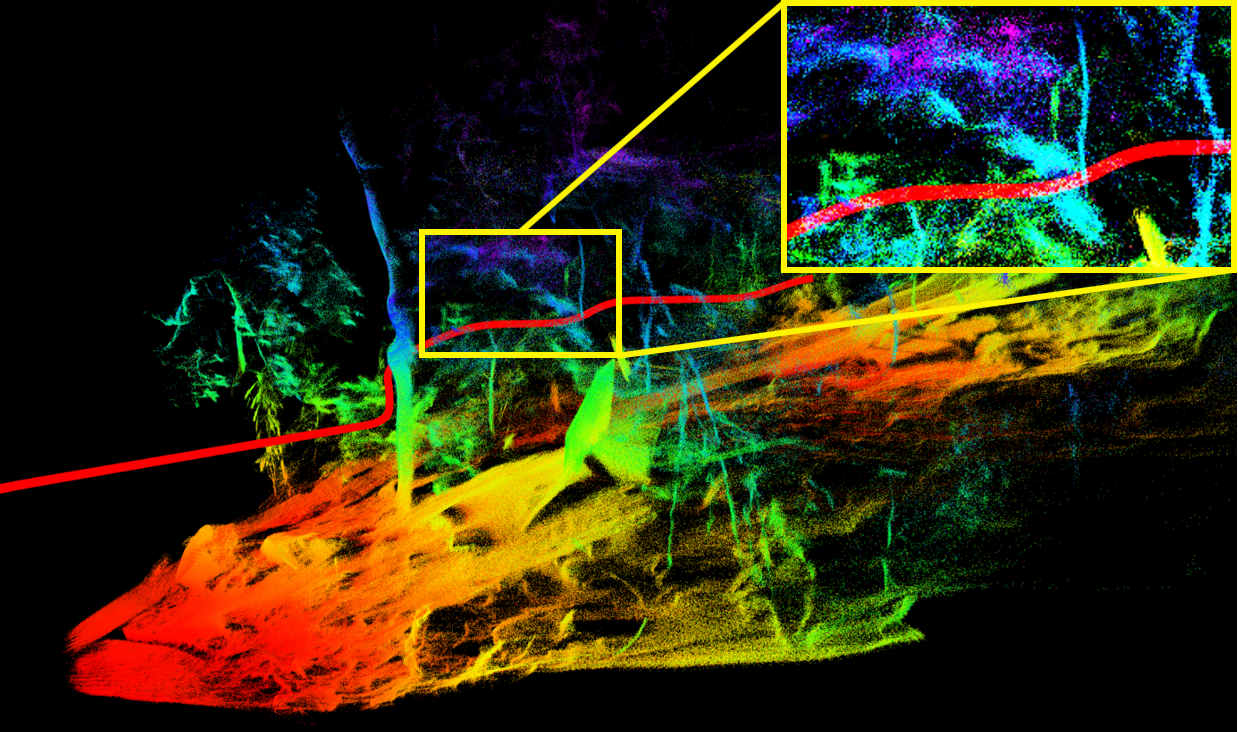}
\caption{Global point cloud map of the second outdoor environment, with leafy trees, stones and a water pipe. The upper right corner shows the leaves around the trajectory.}
\label{outdoor_environment2_figure}
\vspace{-8mm}
\end{figure}

The second outdoor environment is a very complex hillside full of cluttered tree crowns, stones, water pipes, and with significant height variation, as shown in Fig. \ref{outdoor_environment2_figure}. We set the target point of the UAV to be 18 meters away from its takeoff position. Our method is able to find a safe path through cluttered tree crowns and flies to the target point. A part of the path is in a narrow corridor as shown in the upper right corner of Fig. \ref{outdoor_environment2_figure} but the UAV still succeeds.

\subsection{Breakdown of running time}
For a fully autonomous UAV system, a physical limitation that cannot be ignored is the limitation of computing resources. It is impossible and costly for a small quadrotor to carry a desktop-grade PC due to the weight constraint. Therefore, the algorithm speed optimization on the onboard computer is essential for UAVs. In our systems, we utilize several methods to speed up our algorithms.

\begin{figure}[h!]
\centering
\includegraphics[width=1.0\linewidth]{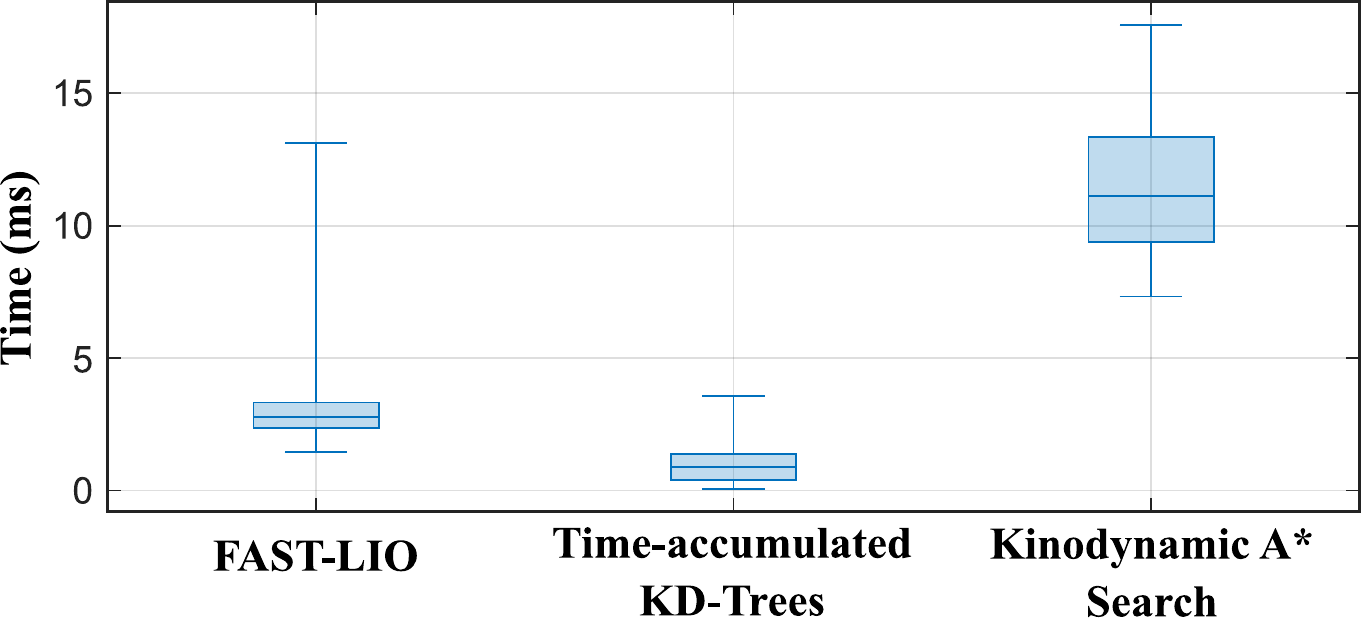}
\caption{Running time statistics of each module for a 7 meters away target in the first outdoor forest environment.}
\label{running_time}
\end{figure}

In the motion planning module, we carefully design the time-accumulated KD-Tree and make the processing time of a new scan of point cloud data less than 5ms, which is described in detail in Section \ref{section:time_accumulated_kdtree}. The resultant computation time of building time-accumulated KD-Tree differs periodically from 0ms to 4ms due to the change of points size in the building process. The computing time of kinodynamic A* search depends on the distance from the target point. For a 7 meters away target in the first outdoor forest environment, it takes about 12ms to generate a trajectory. The running time breakdown is detailed in Fig. \ref{running_time}.

In general, our system achieves real-time capability and is able to avoid dynamic small obstacles. For a 7 meter away target, our system took an average of 17ms, including 4ms (average) for state estimation and mapping, 1ms for updating time-accumulated KD-Trees, and 12ms for kinodynamic A* search. Note that our re-planning is triggered only when the current trajectory collides with an obstacle instead of re-planning at a fixed rate, hence the worst case in Fig. \ref{running_time} rarely occurs.

\begin{figure}[t]
\centering
\includegraphics[width=\linewidth]{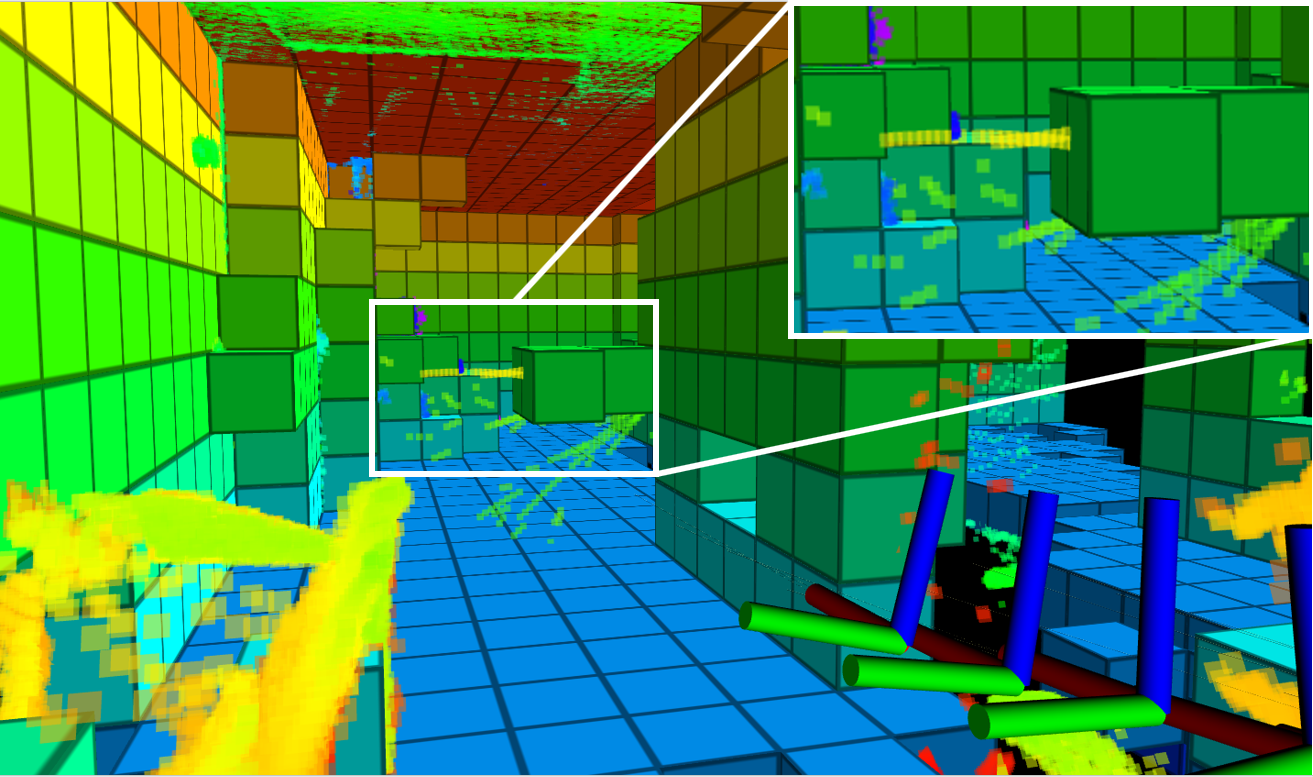}
\caption{Comparision of Octomap and point cloud map. The resolution of Octomap is 0.3m.}
\label{comparision_mapping}
\vspace{-3mm}
\end{figure}

\subsection{Comparison of mapping method}
We finally study the applicability of Octomap  \cite{hornung_octomap_2013} in representing small dynamic objects. Octomap is a popular occupancy grid map that has been widely used in many existing works \cite{gao2020teach,sikang_liu_high_2016} and also constitutes the basis for ESDF maps \cite{oleynikova_voxblox_2017,han2019fiesta}. Taking the indoor office corridor as an example, we build an Octomap at 30cm resolution, as shown in Fig. \ref{comparision_mapping}. In the figure, we also display the point cloud for comparison. It is seen that at this resolution, only the root part of the 2cm bar is detected due to the larger human arm holding the bar, while the main body of the bar, although being evident in the point cloud map, is not detected at all. The reason is that the grid containing the narrow bar has rays from the background objects. When the bar is very narrow, the grid contains many more rays from the background objects than rays from the bar itself. As a result, the ray-cast method would quickly lower the occupancy probability of the grid because it is ``seen through" by many background points' rays. This situation is even worse when the small bar is moving. We have also tested higher resolution of the Octomap including 20cm, 10cm, and 5cm, but none of them reliably detect the bar.

Updating the Octomap also takes significant time because, for a new point, all grids on the ray need to be updated. At a resolution of 30cm and a scan rate of 50Hz, the incremental update of Octomap takes 5-8ms time. This update time will dramatically increase when using a higher resolution or in outdoor open areas. In this case, new points are tens of or hundreds of meters away and for each new point, many grids on the ray need to be updated. On the other hand, our methods directly plan on the point cloud, the time for building (or updating) KD-trees depends on the number of points (which is nearly constant in every new scan) instead of the points' locations.

\section{Conclusion And Future Work}

This paper proposes an autonomous UAV system where all the state estimation, mapping, and trajectory planning are performed onboard based solely on lidar and inertial measurements. The design of the hardware and software system is presented. Extensive indoor and outdoor experiments are conducted. Results show that the proposed system is able to fly safely in cluttered environments while avoiding dynamic small objects. 

The proposed system uses a Kinodynamic A* (re-) planning method, where the control (i.e., acceleration) space is discretized to produce motion primitives. The searched trajectory has a discontinuous acceleration trajectory, preventing the UAV from high flying speeds requiring accurate trajectory tracking. The small number (i.e., 27) of control-space motion primitives also limits the UAV from operating in an even tighter space or higher re-planning rate. In the future, we would like to improve the re-planning module to enable more challenging environments and higher flying speeds.

\bibliography{IEEEabrv,Bibliography}

\vfill

\end{document}